\documentclass[letterpaper, 10 pt, conference]{ieeeconf}  

\IEEEoverridecommandlockouts                              

\usepackage{graphics} 
\usepackage{graphicx}

\usepackage{epsfig} 
\usepackage{amsmath} 
\usepackage{amssymb}  

\usepackage{algorithm}
\usepackage[noend]{algpseudocode} 
\algnewcommand\AAND{\textbf{ and }}
\algnewcommand\Or{\textbf{ or }}

\usepackage{color}
\usepackage{citesort}
\usepackage{url}
\usepackage{breakurl}
\usepackage[breaklinks]{hyperref}

\DeclareMathAlphabet{\pazocal}{OMS}{zplm}{m}{n}

\newcommand{\Ys}{\pazocal{Y}}

\newcommand{\Ebb}{\mathbb{E}}
\newcommand{\Sbb}{\mathbb{S}}
\newcommand{\Mbb}{\mathbb{M}}
\newcommand{\Gbb}{\mathbb{G}}
\newcommand{\Vbb}{\mathbb{V}}

\def \*#1 {mathbf{#1}}
\def \@#1 {\mathbb{#1}}

\newtheorem{definition}{Definition}

\DeclareMathAlphabet{\mathpzc}{OT1}{pzc}{m}{it}

\usepackage{array}
\newcolumntype{C}[1]{>{\centering\arraybackslash}p{#1}}
\newcolumntype{M}[1]{>{\raggedright\arraybackslash}p{#1}}

\usepackage{array} 
\newcolumntype{L}[1]{>{\raggedright\let\newline\\\arraybackslash\hspace{0pt}}m{#1}}	
\newcolumntype{S}[1]{>{\centering\let\newline\\\arraybackslash\hspace{0pt}}m{#1}}
\newcolumntype{R}[1]{>{\raggedleft\let\newline\\\arraybackslash\hspace{0pt}}m{#1}}

\newtheorem{problem}{Problem}

\usepackage[nolist,nohyperlinks]{acronym}
\acrodef{swap}[SWAP]{Semantics-aWAre exploration and inspection Planner}
\acrodef{fov}[FoV]{Field of View}
\acrodef{tsp}[TSP]{Traveling Salesman Problem}

\makeatletter
\renewcommand*{\@opargbegintheorem}[3]{\trivlist
  \item[\hskip \labelsep{\itshape #1\ #2}] \textit{(#3)}\ }
\makeatother

\title{\LARGE \bf
Semantics-aware Exploration and Inspection Path Planning
}

\author{Mihir Dharmadhikari and Kostas Alexis
\thanks{This material was supported by the Research Council of Norway under project SENTIENT (NO-321435).}
\thanks{The authors are with the Autonomous Robots Lab, Norwegian University of Science and Technology (NTNU), O. S. Bragstads Plass 2D, 7034, Trondheim, Norway {\tt\small mihir.dharmadhikari@ntnu.no}}
}

\begin{document}

\maketitle
\thispagestyle{empty}
\pagestyle{empty}

\begin{abstract}
This paper contributes a novel strategy for semantics-aware autonomous exploration and inspection path planning. Attuned to the fact that environments that need to be explored often involve a sparse set of semantic entities of particular interest, the proposed method offers volumetric exploration combined with two new planning behaviors that together ensure that a complete mesh model is reconstructed for each semantic, while its surfaces are observed at appropriate resolution and through suitable viewing angles. Evaluated in extensive simulation studies and experimental results using a flying robot, the planner delivers efficient combined exploration and high-fidelity inspection planning that is focused on the semantics of interest. Comparisons against relevant methods of the state-of-the-art are further presented. 
\end{abstract}

\section{INTRODUCTION}\label{sec:intro}
Robotic systems have long been utilized for remote sensing and inspection tasks~\cite{sa2014vertical,gehring2019anymal,caprari2012highly,chan2015towards,BABOOMS_ICRA_15,hollinger2012uncertainty}. Flying or ground robots, for example, are actively utilized to explore and inspect industrial facilities~\cite{sa2014vertical,gehring2019anymal,caprari2012highly,chan2015towards,BABOOMS_ICRA_15} or even demanding subterranean settings~\cite{CERBERUS_SCIENCE_2022,CERBERUS_WINS_FR2022submission,GBPLANNER2COHORT_ICRA_2022,GBPLANNER_JFR_2020,hudson2021heterogeneous,agha2021nebula,rouvcek2019darpa,explorer_phase_i_ii}. Building upon this success, the research community and the industry are currently actively looking towards means to completely automate the process of building and maintaining accurate ``digital twins''~\cite{tao2018digital} of the facilities of interest. In this framework, explicit focus on specific semantics of interest is of paramount importance, a task assisted by the progress in semantic segmentation and mapping~\cite{schmid2022panoptic,hu2020randla,garcia2017review,grinvald2019volumetric,mccormac2018fusionpp,rosinol2020kimera}. Within a large-scale industrial facility, or other environment of interest, the vast majority of the surfaces may not represent informative or generally significant regions. It is in fact mostly specific structures that require precise and comprehensive monitoring. Accordingly, traditional methods on exploration or coverage path planning are not tailored to efficiently undertake the task as they are agnostic to semantics, rendering them inefficient and unable to deliver the necessary inspection behavior at scale. 


\begin{figure}[ht]
\centering
    \includegraphics[width=0.99\columnwidth]{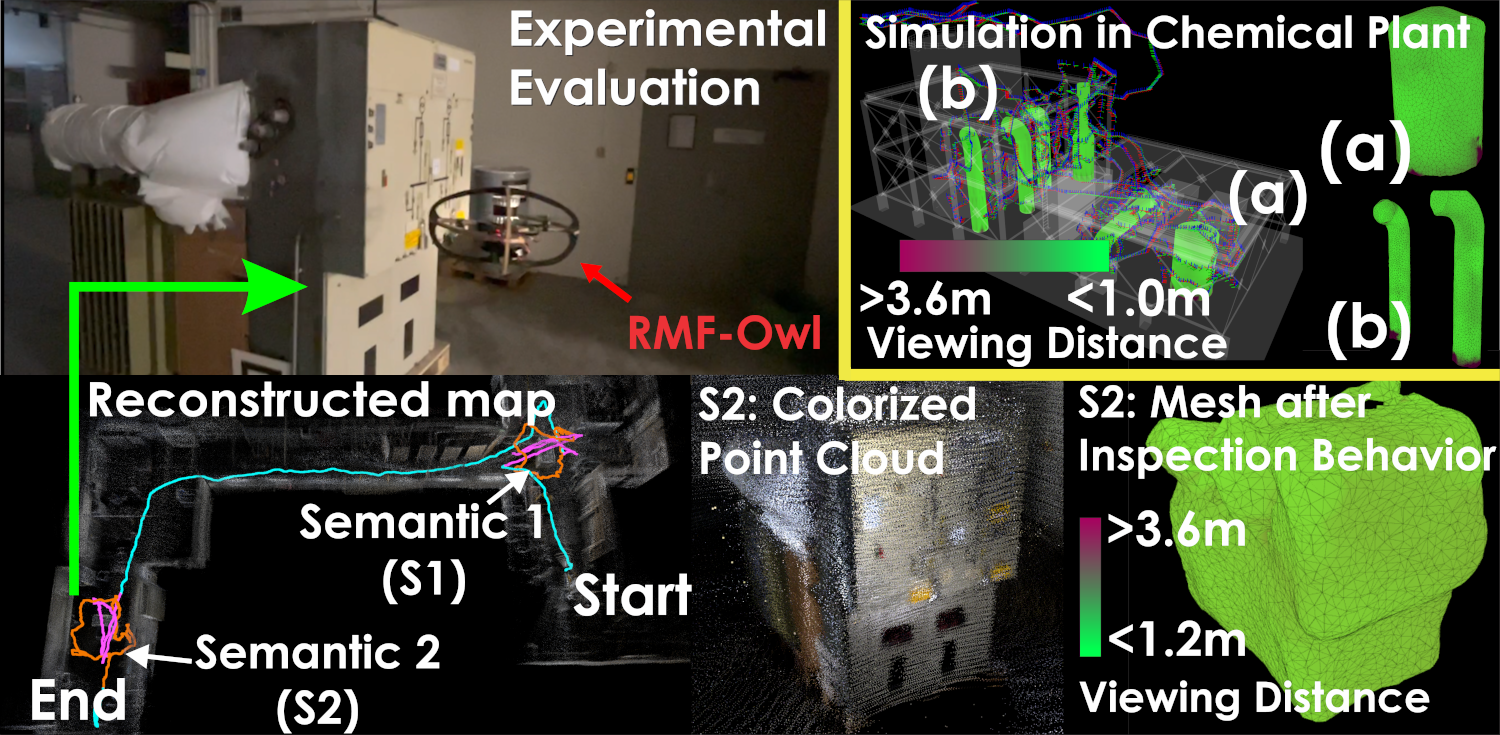}
\vspace{-4ex}
\caption{Instance of an autonomous semantics-aware exploration and inspection mission in an environment with two sparsely distributed semantics (S1, S2). The path traversed by the robot and the reconstructed meshes colorized by camera viewing distance are shown. Furthermore, a simulation run inside a chemical plant is depicted. }
\label{fig:swap_intro}
\vspace{-2ex}
\end{figure}

Motivated by the above, in this work we present a novel \ac{swap} that is tailored to environments with sparse semantics, assumes no prior map information, and can ensure as complete mesh reconstruction and surface inspection as possible at a set minimum resolution. The method is not exploiting merely a volumetric representation~\cite{voxblox,hornung13auro}, as commonly done in exploration planners~\cite{GBPLANNER_JFR_2020,NBVP_ICRA_16}, and does not make the often unrealistic assumption of having access to an a priori available high-resolution overall mesh representation of the environment. Instead, \ac{swap} combines a volumetric map with a disconnected set of meshes focused only on the semantics of interest which are also utilized to perform calculations as to coverage completeness or inspection fidelity in order to best plan its actions. Importantly, the method ensures autonomous exploration within unknown maps combined with complete and selectively detailed semantic inspection by means of transitioning among a set of behaviors as opposed to merely combining distinct objectives in an additive manner that would be highly susceptible to tuning. In particular, \ac{swap} offers efficient volumetric exploration using a range sensor (e.g., $3\textrm{D}$ LiDAR), while at the same time for the detected semantics of interest it enables capturing the further point clouds necessary to derive a complete mesh model for each, and ensures that their surfaces are viewed by a --possibly separate-- camera at sufficient resolution and through suitable viewing angles.  

To verify the proposed solution for semantics-aware exploration and inspection path planning, a set of simulation and experimental studies are presented using flying robots, including comparison against a state-of-the-art exploration planner that does not explicitly consider semantics. We demonstrate that \ac{swap} presents significant advantages for combined exploration and inspection tasks involving a sparse distribution of semantics, the mesh models of which have to be reconstructed and their surfaces be captured by a camera with image quality guarantees. Accordingly, we believe that the method is tailored to be used in complex facilities for which digital twins have to be built and then maintained. 


In the remaining paper, Section~\ref{sec:related} presents related work, followed by the problem statement in~\ref{sec:probstat}. The proposed approach is detailed in Section~\ref{sec:approach}, with evaluation studies in Section~\ref{sec:evaluation} and conclusions in Section~\ref{sec:concl}.

\section{RELATED WORK}\label{sec:related}
This work on semantics-aware exploration and inspection planning has relevance to a host of contributions that relate to its functionality. With respect to its elementary behavior on exploration planning it builds upon the relevant literature~\cite{NBVP_ICRA_16,bircher2016receding,GBPLANNER2COHORT_ICRA_2022,GBPLANNER_JFR_2020,yamauchi1997frontier,cao2021tare,best2022resilient} and especially our prior work on graph-based exploration~\cite{GBPLANNER2COHORT_ICRA_2022,GBPLANNER_JFR_2020}. Similarly, regarding its further core behavior of inspection planning over a reconstructed mesh it relates to works on coverage path planning~\cite{SIP_AURO_2015,bircher_robotica,soltero2014decentralized,choset1997coverage,galceran2013survey,hover2012advanced}. 

With respect to the key innovation in semantics-aware planning, this work relates to two sets of contributions. First, it partially relates to works that on one hand only focus on planning for safe navigation among predefined waypoints but also simultaneously consider the semantics in the environment~\cite{bartolomei2021semantic,bartolomei2020perception,ryll2020semantic}. Second, and most importantly, the proposed method relates to the narrow niche of contributions that aim to co-optimize exploration and mapping of objects/semantics of interest~\cite{dang2018autonomous,koch2019automatic,achatpath,suriani2021s}. Our earlier work in~\cite{dang2018autonomous} performs autonomous exploration but also allows for small deviations from the exploration path in order to improve the resolution of observation of detected objects based on a volumetric map. The contribution in~\cite{koch2019automatic} considers an a priori known map and formulates the planning problem as an orienteering one further considering the desired reconstructed model resolution. The method in~\cite{achatpath} uses a 3-layer semantic map involving a costmap, a classical exploration grid and a binary grid that monitors the observation over specific semantic classes and derives semantics-aware adaptations of A$^\star$, transition-based RRT, and a shortcut algorithm which are then tested on a combined exploration and observation task. The authors in~\cite{suriani2021s} use object detection and explicitly combine semantic information with exploration planning techniques to improve the quality of the $3\textrm{D}$ reconstruction. 


Compared to the existing literature, \ac{swap} contributes multiple innovations and key features. First, it makes no assumption about the shape, size and location of the semantics or any prior map knowledge. Second, it combines exploration and semantic inspection in a principled manner through explicit behaviors as opposed to techniques that merely aggregate goals in an additive objective. Accordingly, it ensures complete semantic mapping using a range sensor (no gaps in the reconstructed mesh) and inspection with a camera at any desired resolution up to what is possible through collision-free configurations. Finally, it realizes a task-specific multi-facet environment representation involving a sparse volumetric map, a semantics-specific locally dense mesh representation and a sparse global graph for replanning in the mapped space. 

\section{PROBLEM FORMULATION}\label{sec:probstat}







The overall problem considered in this work is that of combined exploration of a bounded volume $V \subset \mathbb{R}^3$ with a depth sensor $\Ys_D$ and detailed inspection of a set $\Lambda$ of structures of interest within it, called semantics, in terms of reconstructing meshes of their surfaces using segmented data from $\Ys_D$, and scanning them using a visual camera sensor $\Ys_C$. The problem can be cast globally as that of a) determining which parts of the initially unexplored volume $V_{une}\overset{init.}{=} V$ are free $V_{free}\subseteq V$ or occupied $V_{occ}\subseteq V$ b) creating complete mesh reconstructions $\mathcal{M}^j$ of the surfaces $S_\mathbb{S}^j$ of all the semantics $\mathbb{S}^j$, and c) inspecting every face $f$ of every $\mathcal{M}^j$ with $\Ys_C$ at resolution $r$, defined as the number of pixels per unit area, higher than or equal to $r_{min}$ and viewing angle $\theta_I$ with respect to the outward normal $\mathbf{n}_f$ to $f$ less than $\theta_{I,max}$.
The environment is represented as a volume discretized in an occupancy map $\mathbb{M}$ consisting of cubical voxels $m\in\mathbb{M}$ with edge length $\lambda_V$, and a set of surface meshes $\{\mathcal{M}^j\},~ \forall \mathbb{S}^j \in \Lambda$. 
The operation is subject to the robot dynamics and the sensors' visibility constraints based on two possibly distinct frustum models for $\Ys_D,\Ys_C$ defined by the \ac{fov} $[F_h^D, F_v^D],[F_h^C,F_v^C]$ respectively and maximum ranges $d_{\max}^D,d_{\max}^C$. Hence, certain volume and surface cannot be mapped or inspected resulting in residual volume and surface. Given the above, we have the following definition and problem formulation cast globally.

\begin{definition}[Residual Volume and Semantic Surface]\label{def:residualSpace}
 Let $\Xi$ be the simply connected set of collision free configurations and $\bar{\mathcal{V}}_m\subseteq \Xi$ the set of all configurations from which the voxel $m$ can be perceived by $\Ys_D$. Then the residual volume is $V_{res} = \bigcup_{m\in \mathbb{M}} ( m \vert\ \bar{\mathcal{V}}_m = \emptyset )$. Similarly the residual semantic surface $S_{res}$ can be defined as the part of $\{S_{\mathbb{S}^j}\},~ \forall \mathbb{S}^j \in \Lambda$ that cannot be mapped by $\Ys_D$ or seen by $\Ys_C$.
\end{definition}

\begin{problem}[Volumetric Exploration and Semantic Inspection Problem]\label{prob:swapProblem}
 Given a volume $V$ and an initial configuration $\xi_{init} = [x,y,z,\psi] \subset \Xi$ find a collision-free path $\sigma$ that when traversed by the robot leads to a) identifying $V_{free}$ and $V_{occ}$ based on $\Ys_D$ b) complete mesh reconstructions $\{\mathcal{M}^j\}$ of $\{S_{\mathbb{S}^j}\},~ \forall \mathbb{S}^j \in \Lambda$, and c) inspection of every face $f$ of $\mathcal{M}^j$ with $\Ys_C$ at the required resolution and viewing angle. 
\end{problem}

\section{PROPOSED APPROACH}\label{sec:approach}
\begin{figure}[ht]
\centering
    \includegraphics[width=0.99\columnwidth]{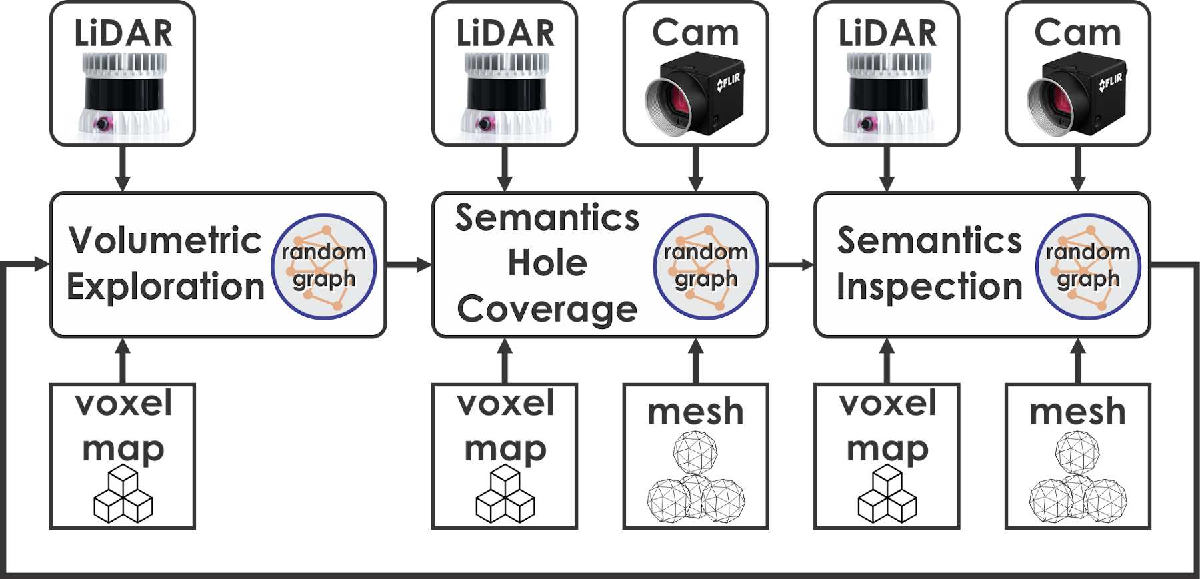}
\vspace{-3ex}
\caption{The behaviors of the semantics-aware exploration and inspection planner. SWAP transitions between autonomous volumetric exploration, semantics mesh hole coverage and semantics inspection considering onboard depth and camera sensors and exploiting a rich environment representation combining volumetric maps and a set of semantic-specific meshes. For all planning stages, a random graph data structure is used for path planning.}
\label{fig:swapbehaviors}
\vspace{-2ex}
\end{figure}

The proposed semantics-aware exploration and inspection planner (SWAP) utilizes three planning behaviors, namely Volumetric Exploration, Semantics Hole Coverage, and Semantics Inspection (Figure~\ref{fig:swapbehaviors}), for efficient exploration of the unknown environment combined with targeted inspection of the surfaces of the semantics of interest. The planner starts in the exploration mode to efficiently explore and volumetrically map the unknown environment. During exploration, possible semantic detections --defining which points of the $\Ys_D$ measurements belong to a certain semantic-- are used to generate meshes as described in Section~\ref{subsec:env_rep}. The planner performs exploration for an allotted time $T_e$. After that, it switches to the hole coverage mode and identifies the holes possibly left in the semantic meshes. The semantic closest to the robot is selected first and viewpoints considering the onboard depth sensor $\Ys_D$ are planned to iteratively fill these holes in meshes of all of the detected semantics as described in Section~\ref{subsec:hole_cov}. Once all holes in the meshes of all the semantics detected so far are filled, no more collision-free configurations can be found to fill the remaining holes, or an allotted time for that semantic has expired, the planner switches to the inspection mode in order to provide detailed inspection of the semantics of interest considering the onboard camera sensor $\Ys_C$ driven by certain metrics. Once this process is also complete within the currently explored map, the method switches back to exploration mode to explore further and where and when needed switch to the semantics-driven behaviors. 

\subsection{Environment Representation} \label{subsec:env_rep}
For efficient exploration and semantic inspection, SWAP utilizes a dual representation of the environment consisting of a volumetric map $\mathbb{M}$ and a set $\mathbb{D}$ of data structures $D^j$ storing a reconstructed point cloud $P_R^j$, a surface mesh $\mathcal{M}^j$, and corresponding raw sensor data for each semantic $\mathbb{S}^j \in \Lambda$. In this work we use Voxblox~\cite{voxblox} as the volumetric mapping framework for collision-free navigation and volumetric calculations.
Along with the volumetric map, we generate surface meshes for each detected semantic in the environment using semantically segmented $\Ys_D$ data. In this work, we assume that a semantic segmentation module providing segmented point cloud exists. However, the method does not make any other assumptions about the size, shape, or location of semantics in the environment.
From every segmented point cloud $P_{seg}$, the points $P_{seg}^j$ belonging to semantic $\mathbb{S}^j$ are appended to $P_R^j$. At a fixed temporal frequency, $P_R^j$ is subsampled, and the surface mesh $\mathcal{M}^j$ is constructed using the Advancing Front Surface Reconstruction algorithm~\cite{advancing_front}. The faces of $\mathcal{M}^j$ that are observed by $\Ys_C$ at a desired resolution $r_{\min}$ and viewing angle $\theta_{I,\max}$ are further marked.


\subsection{Volumetric Exploration}\label{subsec:exploration}


\ac{swap} implements its autonomous exploration functionality by interfacing our previous and open-sourced work on graph-based exploration (GBPlanner)~\cite{GBPLANNER2COHORT_ICRA_2022,GBPLANNER_JFR_2020,CERBERUS_SCIENCE_2022}. The method, verified extensively in subterranean and industrial environments, offers efficient exploration within a volume of set bounds assuming no prior map knowledge. It operates over a volumetric representation of the environment based on~\cite{voxblox} and functions in a bifurcated architecture of local- and global path planning. At the local stage the method exploits a dense random graph $\Gbb^L_E$ around the robot to identify collision-free paths maximizing volumetric exploration. Simultaneously, as such local steps take place, the algorithm builds a sparse global graph $\Gbb^G_E$, used by the global stage that is invoked when local exploration reports inability to find a path of significant gain or when the robot approaches its endurance limits. Accordingly, the method offers re-positioning to previously detected unexploited frontiers of the exploration space or timely auto-homing. In SWAP, the autonomous exploration behavior is invoked for $T_e$ seconds, before the system possibly switches to its semantically-driven behaviors.

\begin{figure}[ht]
\centering
    \includegraphics[width=0.9\columnwidth]{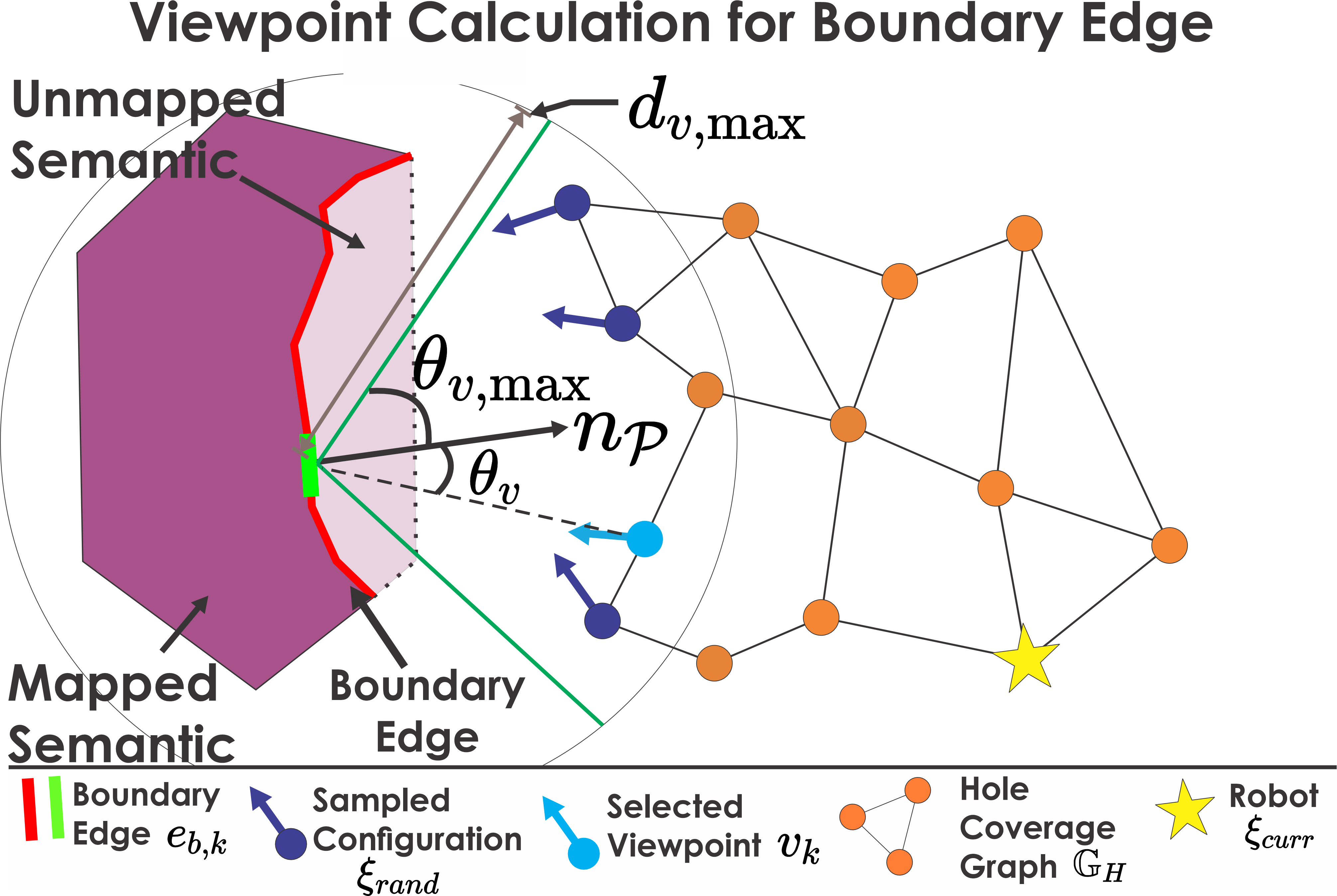}
\vspace{-1ex}
\caption{Viewpoint generation procedure for a boundary edge of a hole during Semantics Hole Coverage mode. Robot configurations are sampled in a spherical coordinate system centered at the edge. The configurations that respect the viewing angle criteria (as described in Section~\ref{subsec:hole_cov}) and can be connected to the hole coverage graph are extracted, and the configuration with the least viewing angle is selected as the viewpoint.}
\label{fig:hole_viewpoint_sampling}
\vspace{-1ex}
\end{figure}

\subsection{Semantics Hole Coverage}\label{subsec:hole_cov}


After performing exploration for $T_e$ seconds, the planner identifies the set of semantics, $\Lambda_H$, whose reconstructed meshes contain holes, and attempts to plan viewpoints to collect measurements to fill them considering one semantic at a time. The iterative procedure for hole coverage of one semantic is described in Algorithm~\ref{alg:hole_coverage}.

The planner first selects the semantic $\Sbb_{C,H} \in \Lambda_H$ closest to the robot. For the mesh $\mathcal{M}_{\mathbb{S}_{C,H}}$ of semantic $\Sbb_{C,H}$ the planner first identifies the set $\Ebb_b$ of mesh edges that belong to only one mesh face, called boundary edges $e_{b,k} \in \Ebb_b$, and their corresponding faces called boundary faces $f_{b,k}$.
We treat the boundary edges as frontiers on the mesh that need to be covered iteratively. For each boundary edge $e_{b,k}$ a viewpoint $v_k$ is derived that can view the edge and map the region around it as shown in Figure~\ref{fig:hole_viewpoint_sampling}. Furthermore, the planner maintains a collision-free $3\textrm{D}$ graph, called the hole coverage graph $\Gbb_H$, connecting the current robot configuration $\xi_{curr}$ to all viewpoints $v_k$.

\begin{algorithm}
\caption{Semantics Hole Coverage}\label{alg:hole_coverage}
\begin{algorithmic}[1]
\State $\Lambda_H \gets \mathbf{extractSemanticsWithHoles}(\Lambda)$
\While {$ \Lambda_H \neq \emptyset $}
    \State $\mathbb{S}_{C,H} \gets \mathbf{closestSemantic}(\Lambda_H)$
    \State $\mathbb{E}_b \gets \mathbf{extractBoundaryEdges}(\mathbb{S}_{C,H})$
    \While{$\Ebb_b \neq \emptyset ~ \textrm{and} ~ \mathbf{timeRemaining}=\textrm{True}$}
        \State $W_c \gets \mathbf{calculateViewpoints}(\Ebb_b, \Mbb, \Gbb_H)$
        \If{$W_c = \emptyset$}
            \State exit loop
        \EndIf
        \State $W_s \gets \mathbf{reduceNumberOfViewpoints}(W_c)$
        \State $\sigma_{hc} \gets \mathbf{pathToClosestViewpoint}(W_s, \Gbb_H)$
        \State $\mathbf{executePath}(\sigma_{hc})$
        \State $\mathbf{updateMesh}(\Sbb_{C,H})$
    \EndWhile
    \State $\Lambda_H \gets \Lambda_H \setminus \Sbb_{C,H}$
    \State $\Lambda_H \gets \Lambda_H \bigcup \mathbf{newSemanticsWithHoles}$
\EndWhile
\end{algorithmic}
\end{algorithm}

To generate the viewpoint for an edge, the planner randomly samples collision-free robot configurations $\xi_{rand}$, with $\Ys_D$ pointing towards $e_{b,k}$, in a spherical coordinate system centered at the midpoint of the edge $e_{b,k}$ with maximum radial distance $d_{v,\max}$.
The graph $\Gbb_H$ is extended towards $\xi_{rand}$ and the configurations that cannot be connected to $\Gbb_H$ are discarded. Out of the remaining configurations, the one whose viewing angle $\theta_v$, defined as the angle between the tangential vector $\mathbf{n}_\mathcal{P}$ to $f_{b,k}$ and the vector from the center of $e_{b,k}$ to $\xi_{rand}$ (Figure~\ref{fig:hole_viewpoint_sampling}), is lowest and less than a threshold $\theta_{v,\max}$ is selected as the viewpoint for that edge and added to the set $\mathcal{W}_c$ of candidate viewpoints.
To reduce the number of viewpoints, a viewpoint $v_l$ is selected from $\mathcal{W}_c$ and added to the set $\mathcal{W}_s$ of selected viewpoints. All boundary edges within a distance $d_{v,\max}$ from it for which $\theta_v$ of $v_l$ is less than $\theta_{v,\max}$ are extracted, and the corresponding viewpoints are removed from $\mathcal{W}_c$. This procedure is repeated till $\mathcal{W}_c$ is empty.
The path $\sigma_{hc}$ to the viewpoint $v_{close} \in \mathcal{W}_s$ having the least path length from $\xi_{curr}$ along $\Gbb_H$ is calculated and executed by the robot. This procedure is repeated until either no more boundary edges exist or no admissible viewpoints can be found for any boundary edge. Additionally, another threshold is introduced for the size of the hole removing small gaps, as well as a maximum time $T_{hc}$ is allotted for hole coverage of each semantic. Once the hole coverage process is finalized for one detected semantic, the planner proceeds to the next detected semantic before eventually switching to inspection mode.

\begin{figure}[ht]
\centering
    \includegraphics[width=0.9\columnwidth]{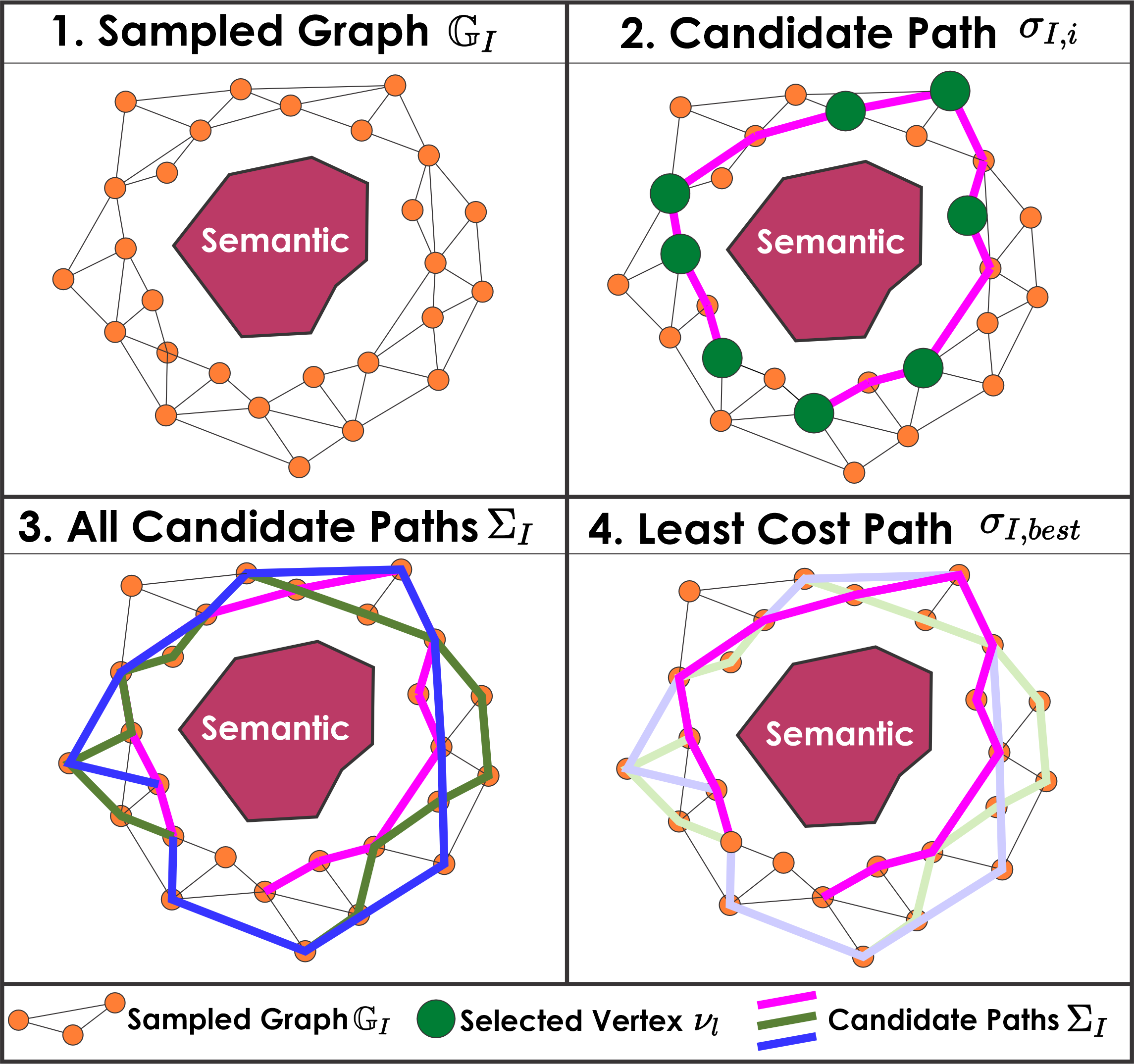}
\vspace{-1ex}
\caption{Illustration of the steps involved in the Semantics Inspection mode of SWAP. A collision-free graph is built around the semantic. $k$ candidate paths providing complete coverage are calculated from the graph, and the path having the least cost is executed by the robot.}
\label{fig:inspection_planning}
\vspace{-1ex}
\end{figure}

\begin{figure*}[ht!]
\centering
\includegraphics[width=\textwidth]{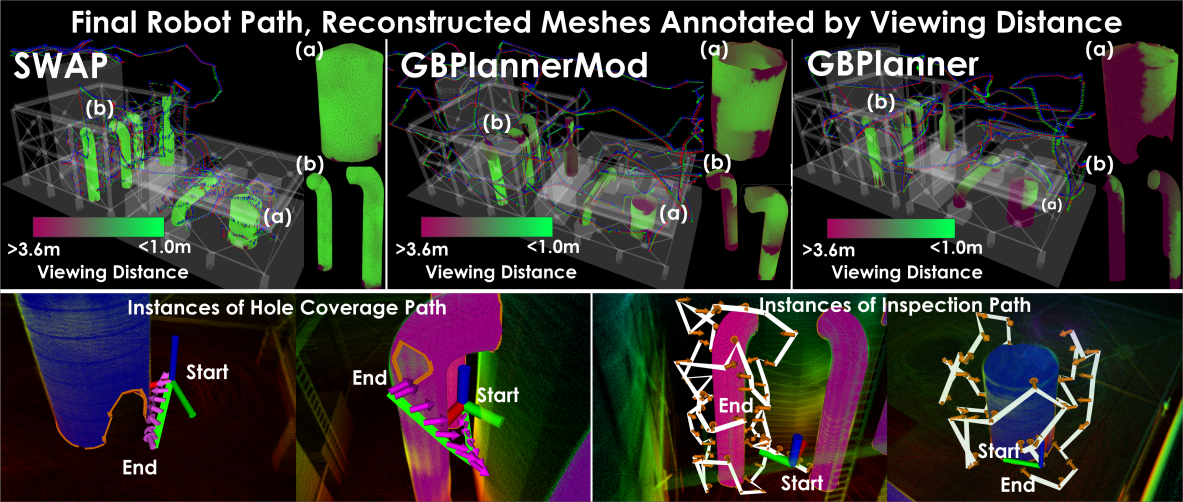}
\caption{This figure shows the results of the simulation conducted in the model of a chemical plant. The figures on the top show the overall path traversed by the robot along with the semantic mesh reconstructions annotated with the distance from which the faces of the meshes were viewed for all three planners. As seen from the mesh reconstructions, the proposed planner is able to provide more complete semantic meshes with near complete visual coverage than the state-of-the-art. The bottom row shows instances of hole coverage and inspection paths.}
\label{fig:simulation}
\vspace{-2ex}
\end{figure*}

\subsection{Semantics Inspection}\label{subsec:inspection}

For the reconstructed mesh of each semantic $\Sbb_I$ in the set of detected and partially inspected semantics $\Lambda_I$, the inspection mode of the planner finds an efficient path such that every face of the semantic mesh is inspected from at least one configuration in that path at the desired resolution and viewing angle. 
A face $f$ is said to have been inspected from a configuration $\xi$ if a) the centroid $\mu_f$ of $f$ lies within the modeled sensor $\Ys_C$ frustum, b) the distance $d_f$ from $\xi$ to $\mu_f$ is within a threshold distance $l_{max}$ (calculated based on $r_{min}$), and c) the angle $\theta_I$ between the vector from $\mu_f$ to $\xi$ and the outward normal $\mathbf{n}_f$ is less than a set limit $\theta_{I,max}$. 
\begin{algorithm}
\caption{Semantics Inspection}\label{alg:inspection}
\begin{algorithmic}[1]
\State $\Lambda_I \gets \mathbf{extractUninspectedSemantics}(\Lambda)$
\While {$ \Lambda_I \neq \emptyset ~\textrm{and}~  \Lambda_H = \emptyset$}
    \State $\mathbb{S}_{C,I} \gets \mathbf{closestSemantic}(\Lambda_I)$
    \State $\Gbb_I \gets \mathbf{buildGraph}(\Sbb_{C,I}, \Mbb)$
    \State $c_b \gets \infty, ~ \sigma_{I,best} \gets \emptyset$
    \For {$\textit{i = 1 to k}$}
        \State $\Vbb_c \gets \mathbf{selectCompleteCoverageVertices}()$
        \State $\sigma_{I,i} \gets \mathbf{solveTSP}(\Vbb_c)$
        \If{$\mathbf{cost}(\sigma_{I,i}) < c_b$}
            \State $c_b \gets \mathbf{cost}(\sigma_{I,i}), ~ \sigma_{I,best} \gets \sigma_{I,i}$
        \EndIf
    \EndFor
    \State $\Lambda_I \gets \Lambda_I \setminus \Sbb_{C,I}$
    \State $\Lambda_H \gets \Lambda_H \bigcup \mathbf{newSemanticsWithHoles}$
\EndWhile
\end{algorithmic}
\end{algorithm}
The inspection planning procedure is detailed in Algorithm \ref{alg:inspection} and illustrations are shown in Figure~\ref{fig:inspection_planning}. The planner first selects the semantic $\Sbb_I \in \Lambda_I$ that is closest to the robot, calculates an oriented bounding box $V_s$ around the considered semantic mesh, and samples a $3\textrm{D}$ collision-free graph $\mathbb{G}_I$. For each vertex $\nu_i$ in $\Gbb_I$, the expected unobserved faces seen by $\Ys_C$ from $\nu_i$ are calculated. The set $L_i$ of faces seen by $\nu_i$ is referred to as the visibility of $\nu_i$. Next, a path that ensures full inspection of the mesh, within the limitation of traveling in collision-free space, needs to be searched from this graph. 
To this end, the planner calculates $k$ candidate paths $\sigma_{I,i} \in \Sigma_I,~i=1...k$ providing complete coverage of the semantic mesh and selects the one with the least cost in terms of the path execution time. Each candidate path is calculated as follows.
All vertices $\nu_i$ in $\mathbb{G}_I$ are sorted by the cardinality of $L_i$. The top $\eta \%$ of vertices are selected, one vertex $\nu_l$ is chosen from them at random and added to the set $\mathbb{V}_c$ of vertices in the final path. The visibility of the remaining vertices is re-evaluated to account for the overlap with the vertices in $\mathbb{V}_c$. This process is continued till none of the remaining vertices have non-empty visibility. The order in which the vertices in $\mathbb{V}_c$ are to be visited is determined by solving the \ac{tsp}, by means of the Lin-Kernighan-Helsgaun (LKH) heuristic~\cite{helsgaun2000effective}, where the cost of travel between any two vertices is the length of the shortest path along $\mathbb{G}_I$.
Finally, the path $\sigma_{I,best} \in \Sigma_I$ with the lowest execution time cost $c_b$, is selected and executed by the robot.
At the end of execution, if any new semantics $\Lambda_H$ requiring hole coverage are detected, the planner switches to the hole coverage mode, otherwise moves to the next detected semantic, closest to the robot, for inspection.

\section{EVALUATION STUDIES}\label{sec:evaluation}
To evaluate the proposed semantics-aware exploration and inspection planner, both simulation and experimental studies were conducted. 
The computation times of various steps involved as well as the parameters used in both are presented in Tables~\ref{tab:comp_times} and \ref{tab:params} respectively.

\subsection{Simulation Studies}

We present a simulation study for evaluating the planner inside a model of a chemical plant of size $44 \times 28 \times 20\textrm{m}$ involving six semantics of interest. The simulation study utilized the Gazebo simulator~\cite{GazeboIgn} with a model of the RMF-Owl~\cite{rmfowl,RMFSimModel} aerial robot with dimensions $0.38 \times 0.38 \times 0.24\textrm{m}$ carrying a $3\textrm{D}$ LiDAR sensor as $\Ys_D$ with $[F^D_h,F^D_v] = [360,90]^{\circ}$, $d^D_{\max} = 50\textrm{m}$, and a color camera as $\Ys_C$ having $[F^C_h,F^C_v] = [120,90]^{\circ}$, $d^C_{\max} = 7\textrm{m}$. We utilize the semantic segmentation camera from the simulator to get segmented data. The simulations were conducted on a laptop with an Intel Core i9-10885H CPU.

The performance of the planner was compared against our previous Graph-based Exploration Planner (GBPlanner)~\cite{GBPLANNER_JFR_2020,GBPLANNER2COHORT_ICRA_2022,CERBERUS_SCIENCE_2022} in both its original form (purely volumetric exploration given a depth sensor $\Ys_D$) and with a modified objective for surface coverage. The modified method (hereafter referred to as GBPlannerMod) annotates the mapped voxels that are also seen by $\Ys_C$ in the volumetric map. The method samples a random graph as GBPlanner, and the information gain for each vertex $\nu_i$ is defined as the number of unseen surface voxels inside $\Ys_C$ if the robot were at the configuration $\xi_i$ corresponding to $\nu_i$. An unseen surface voxel is defined as an unknown voxel or an occupied voxel, that is mapped by $\Ys_D$ but not seen by $\Ys_C$, both neighboring a free voxel. Using this new information gain formulation, the path is selected in a manner identical to GBPlanner.

\begin{figure}[H]
\centering
    \includegraphics[width=0.8\columnwidth]{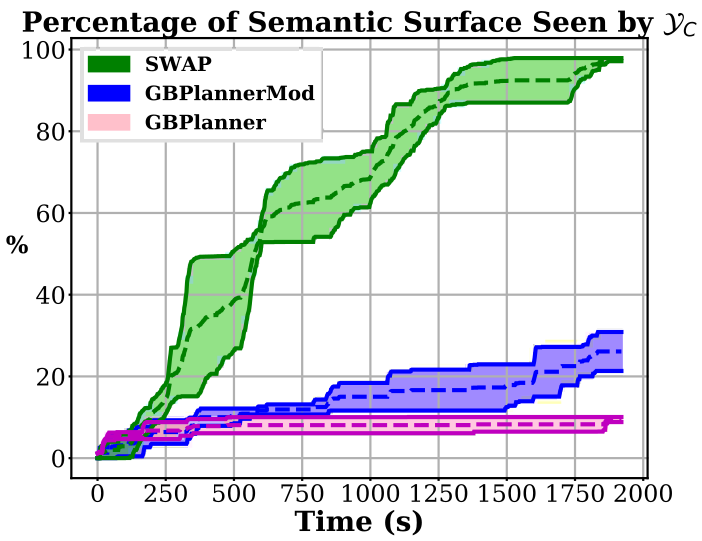}
\vspace{-2ex}
\caption{This plot shows the percentage of the cumulative surface of all semantics seen by $\Ys_C$ over time by all three methods for the conducted simulation study. Each planner was run five times. The targeted inspection of SWAP enables it to significantly outperform the other methods.}
\label{fig:sim_plots}
\vspace{-1ex}
\end{figure}

\begin{figure*}[ht!]
\centering
\includegraphics[width=\textwidth]{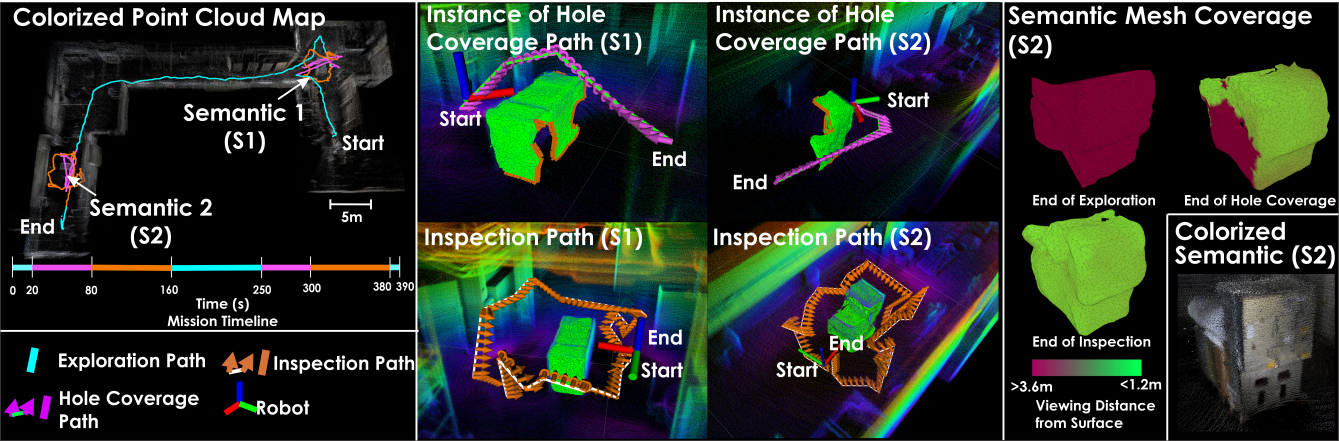}
\caption{The method was verified experimentally by deploying onboard the RMF-Owl aerial robot inside the basement of NTNU's Elektro building. The experiment involved inspection of two machines distributed along three connected corridors connected via two $90^{\circ}$ turns. In this figure, the top right sub-figure shows the overall path traversed by the robot along with a timeline of the mission colored according to the active mode. The sub-figure on the top left presents the mesh reconstruction of one of the semantics with each face colored by the distance from which it was seen by the onboard camera at the end of exploration (up to the beginning of hole coverage for that semantic), hole coverage, and inspection modes. The two sub-figures at the bottom right show the inspection and one of the hole coverage paths for one semantic. Finally the sub-figure on the bottom left shows the point cloud reconstruction of the semantic colorized based on the onboard camera image. The robot was successfully able to explore the environment and inspect both the semantics at the required metrics due to the planner's ability to perform combined exploration and targeted inspection of the semantics.}
\label{fig:experiment}
\vspace{-1ex}
\end{figure*}

A total of five simulation runs of each planner were conducted in the same environment starting from the same location. Each run lasted for $1850\textrm{s}$. The overall paths traversed by the robot, the reconstructed meshes for few of the semantics, and indicative paths from the hole coverage and inspection steps can be seen in Figure~\ref{fig:simulation}. Figure~\ref{fig:sim_plots} shows the semantic visibility comparison between the three methods in terms of the percentage of all semantic surfaces seen by $\Ys_C$, within the required quality metrics, as a function of time. Due to the ability of SWAP to perform targeted inspection of the semantics, it is able to outperform both the other methods. Even GBPlannerMod, with its new formulation for surface inspection, is unable to capture all surfaces of the semantic at the desired quality.

\begin{table}[]
\centering
\vspace{-1ex}
\caption{Computation times for various steps involved.}
\begin{tabular}{|l|l|l|}
\hline
\textbf{Computation Step}           & \textbf{Simulation}               & \textbf{Experiment}            \\ \hline
Volumetric Exploration ($\textrm{s}$)               & $0.648$                & $0.446$                \\ \hline
Semantics Hole Coverage ($\textrm{s}$)             & $0.027$                & $0.121$                \\ \hline
Semantics Inspection ($\textrm{s}$)                & $2.689$                & $1.527$                \\ \hline
Mesh computation ($\textrm{s}$)                    & $0.024$                 & $0.054$                \\ \hline
\end{tabular}
\label{tab:comp_times}
\end{table}

\begin{table}[]
\centering
\vspace{-1ex}
\caption{Parameters used in the simulation and experiment.}
\begin{tabular}{|l|l|l|}
\hline
\textbf{Parameter}          & \textbf{Simulation}              & \textbf{Experiment}            \\ \hline
$\lambda_V~(\textrm{m})$                     & $0.2$          & $0.3$                \\ \hline
$r_{min}~(\textrm{pixels/}\textrm{cm}^2)$   & $20.67$                      & $5.06$                  \\ \hline
$\theta_{I,max}~(\textrm{deg})$                & $45$            & $45$                  \\ \hline
$\theta_{H,max}~(\textrm{deg})$                & $75$            & $75$                  \\ \hline
$k$                             & $10$                      & $3$                \\ \hline
\end{tabular}
\label{tab:params}
\vspace{-3ex}
\end{table}

\subsection{Experimental Studies}

For experimental verification, \ac{swap} is deployed onboard RMF-Owl~\cite{rmfowl}, a small-sized ($0.38\times0.38\times0.24\textrm{m}$) and lightweight ($1.4\textrm{kg}$) aerial robot with approximately $10\textrm{min}$ of endurance integrating a) a multi-modal sensing suite involving an OUSTER OS0 $3\textrm{D}$ LiDAR with $64$ channels used as $\Ys_D$ ($[F^D_h,F^D_v] = [360,90]^{\circ}$, $d^D_{\max}=20\textrm{m}$), a FLIR Blackfly S 0.4MP visual camera used as $\Ys_C$ ($[F^C_h,F^C_v] = [85,64]^{\circ}$, $d^C_{\max} = 7\textrm{m}$), and an IMU, and b) a Khadas VIM3 Pro Single Board Computer (SBC) incorporating $\times 4$ 2.2Ghz Cortex-A73 cores, paired with $\times2$ 1.8Ghz Cortex-A53 cores implementing an A311D big-little architecture. The robot was developed as part of the activities of Team CERBERUS in the DARPA Subterranean Challenge~\cite{CERBERUS_SCIENCE_2022,CERBERUS_WINS_FR2022submission,CERBERUS_SUBT_PHASE_I_II} and integrates a robust localization and mapping method as presented in~\cite{khattak2020complementary} upon which autonomous path planning can take place. 

The conducted experiment took place in the basement of NTNU's Elektro Building and involved two industrial machines as the semantics of interest distributed along three corridors connected via two $90^{\circ}$ turns. The robot started at the beginning of one of the corridors, in the exploration mode with $T_e = 20\textrm{s}$, and after that switched to the semantic modes. Upon completion, it continued exploration and switched to semantic modes after detecting the second semantic. It is noted that as this work is focused on path planning, the semantic detection and point cloud segmentation is considered given and it in the experiment it was implemented by using an AprilTag unique to each semantic object. Figure~\ref{fig:experiment} presents the result of this mission showing the path traversed by the robot, the mission timeline, instances of hole coverage and inspection paths, the generated mesh annotated with the camera viewing distance, and the colorized reconstructed point cloud of one of the semantics. The average resolution at which the semantics S1 and S2 were viewed was $9.81$ and $14.40$ $\textrm{pixels} / \textrm{cm}^2$.

%

\section{CONCLUSIONS}\label{sec:concl}
A semantics-aware path planner for exploration of an unknown environment combined with mesh reconstruction and inspection of the semantics of interest is presented. The behavior-based approach allows the planner to volumetrically explore, generate complete mesh reconstructions of the semantics, and perform an inspection of their faces given specific image quality metrics without any prior knowledge or assumptions about the environment. Both simulation studies comparing the method with a state-of-the-art exploration planner and an experiment using a flying robot are conducted to verify the new method.

\addtolength{\textheight}{-2cm}   




\bibliographystyle{IEEEtran}
\bibliography{./SemanticPlanner_ICRA_2023}

\end{document}